\documentclass[sn-basic]{sn-jnl}

\usepackage{graphicx}%
\usepackage{multirow}%
\usepackage{amsmath,amssymb,amsfonts}%
\usepackage{amsthm}%
\usepackage{mathrsfs}%
\usepackage[title]{appendix}%
\usepackage{xcolor}%
\usepackage{textcomp}%
\usepackage{manyfoot}%
\usepackage{booktabs}%
\usepackage{algorithm}%
\usepackage{algorithmicx}%
\usepackage{algpseudocode}%
\usepackage{listings}%

\begin{document}

\title[Segmentation of macerated fibers and vessels]{Segmentation and Characterization of Macerated Fibers and Vessels Using Deep Learning}


\author[1,2,3]{\fnm{Saqib} \sur{Qamar}}\email{saqib.qamar@umu.se}

\author[5]{\fnm{Abu Imran} \sur{Baba}}\email{abu.baba@slu.se}

\author*[2,4,5]{\fnm{Stéphane} \sur{Verger}}\email{stephane.verger@umu.se}

\author*[1,2,6]{\fnm{Magnus} \sur{Andersson}}\email{magnus.andersson@umu.se}

\affil[1]{\orgdiv{Department of Physics}, \orgname{Umeå University}, \orgaddress{\city{Umeå}, \postcode{90187}, \country{Sweden}}}

\affil[2]{\orgdiv{Integrated Science Lab, Department of Physics}, \orgname{ Umeå University}, \orgaddress{\city{Umeå}, \postcode{90187}, \country{Sweden}}}

\affil[3]{\orgdiv{Robotics, Perception, and Learning (RPL), Department of Intelligent System}, \orgname{KTH Royal Institute of Technology and Science}, \orgaddress{\city{Stockholm}, \postcode{100 44}, \country{Sweden}}}

\affil[4]{\orgdiv{Umeå Plant Science Centre (UPSC), Department of Forest Genetics and Plant Physiology}, \orgname{Swedish University of Agricultural Sciences}, \orgaddress{\city{Umeå}, \postcode{90183}, \country{Sweden}}}

\affil[5]{\orgdiv{Umeå Plant Science Centre (UPSC)}, Department of Plant Physiology \orgname{Umeå University}, \orgaddress{\city{Umeå}, \postcode{90187}, \country{Sweden}}}

\affil[6]{\orgdiv{Umeå Centre for Microbial Research (UCMR)} \orgname{Umeå University}, \orgaddress{\city{Umeå}, \postcode{90187}, \country{Sweden}}}





\abstract{\textbf{Purpose:} Wood comprises different cell types, such as fibers, tracheids and vessels, defining its properties. Studying cells' shape, size, and arrangement in microscopy images is crucial for understanding wood characteristics. Typically, this involves macerating (soaking) samples in a solution to separate cells, then spreading them on slides for imaging with a microscope that covers a wide area, capturing thousands of cells. However, these cells often cluster and overlap in images, making the segmentation difficult and time-consuming using standard image-processing methods.

\textbf{Results:} In this work, we developed an automatic deep learning segmentation approach that utilizes the one-stage YOLOv8 model for fast and accurate segmentation and characterization of macerated fiber and vessel form aspen trees in microscopy images. The model can analyze 32,640 x 25,920 pixels images and demonstrate effective cell detection and segmentation, achieving a mAP$_{0.5-0.95}$ of 78 \%. To assess the model's robustness, we examined fibers from a genetically modified tree line known for longer fibers. The outcomes were comparable to previous manual measurements. Additionally, we created a user-friendly web application for image analysis and provided the code for use on Google Colab.
\textbf{Conclusion:} By leveraging YOLOv8's advances, this work provides a deep learning solution to enable efficient quantification and analysis of wood cells suitable for practical applications.}

\keywords{Instance segmentation, YOLO, wood, fibers, optical microscopy}



\maketitle

\section{Introduction}\label{sec1}
Wood fibers, including tracheids and other fiber types, are essential components of wood, providing mechanical strength to withstand external mechanical stresses such as wind. Moreover, they are crucial for supporting tree growth in height \cite{gorshkova2012plant}. These fibers have a high economic importance as they are the basic constituent of most wood-derived products. Wood fibers are extracted by pulping, which separates these cells into independent fiber cells. Once separated into individual fibers, they can be reassembled into paper and are increasingly utilized in applications such as bio-composites, smart papers, and new packaging materials \cite{gholamp}. The term "wood fiber" generally refers to the cell type called tracheids in softwood (Gymnosperm) and fibers in hardwood (angiosperm) \cite{wilson1986anatomy}. In hardwood, fibers coexist with vessels and ray cells, which collectively determine wood characteristics. Notably, hardwood fiber cells elongate intrusively at their tips. Fiber cells are derived from the same stem cell progenitors as vessels, which do not elongate intrusively \cite{siedlecka2008pectin}. The ratio of fiber and vessel length can indicate the degree of intrusive growth (elongation), an essential factor in determining wood quality \cite{gholamp}. However, despite its importance, little is known about the development and elongation of these fibers or the genetic factors that regulate their length \cite{majda2021elongation}. With this knowledge, fiber properties could be improved to produce better and stronger wood fiber-based products in more sustainable production systems \cite{thumm2013influence}. As the demand for renewable fiber-based products continues to grow, there is a need to develop robust, high-throughput methods for studying fiber length and characteristics. 

A classical method to study wood fiber length consists of macerating (soaking) wood samples in a solution that separates individual cells \cite{siedlecka2008pectin}. Cells suspended in liquid solutions can be transferred onto microscopy slides for examination. Once prepared, these slides can be observed using a standard light microscope. The images captured can then be analyzed using image processing software like ImageJ, which allows for the manual measurement of individual fiber lengths \cite{schneider2012nih}. This task is, however, highly time-consuming and prone to user bias and errors during the manual measurement step. Overall, this highly limits the throughput of this type of analysis. Alternatively, it is possible to use so-called fiber analyzers. These machines allow the high-throughput image acquisition of fibers floating in a constantly stirred solution, generating high-speed and unbiased measurements \cite{eriksson2000increased}. However, the resolution of data acquisition is often limited, and it can be difficult to differentiate individual fibers from clumps of fibers that have not been properly detached from each other. While this may not be a limitation for industrial applications, it can become limiting when it comes to accurately studying the biology of fibers and their individual length, width, and other morphometric descriptors \cite{lobo2016insight}. This type of equipment can also be very costly and is not widely available to many research labs contrary to light microscopes, even those equipped with a motorized stage.

A more widely accessible solution for fundamental research application is thus to make use of commonly available light microscopes equipped with motorized stages. Those are commonly available in biology research labs and most university core microscopy facilities. The motorized stage can be used to automatically acquire very large fields of view with high resolution, usually within 1-2 minutes per slide. Typically, wood macerates mounted between slides and cover slips allow the capture of hundreds to thousands of fibers and vessels per slide. This high-throughput image acquisition can also, in principle, be coupled with new image processing technologies to automate the time-consuming task of identifying and measuring fiber cells in those high-resolution images. The use of automated microscopy slide scanning and automated image processing can thus largely alleviate the drawback of the light microscopy-based approach compared to fiber analyzers while also delivering much higher quality images for detailed fiber characterization.

\begin{figure*}[htb]
\centering{\includegraphics[clip, trim=0cm 0cm 0cm 0cm, width=1\linewidth]{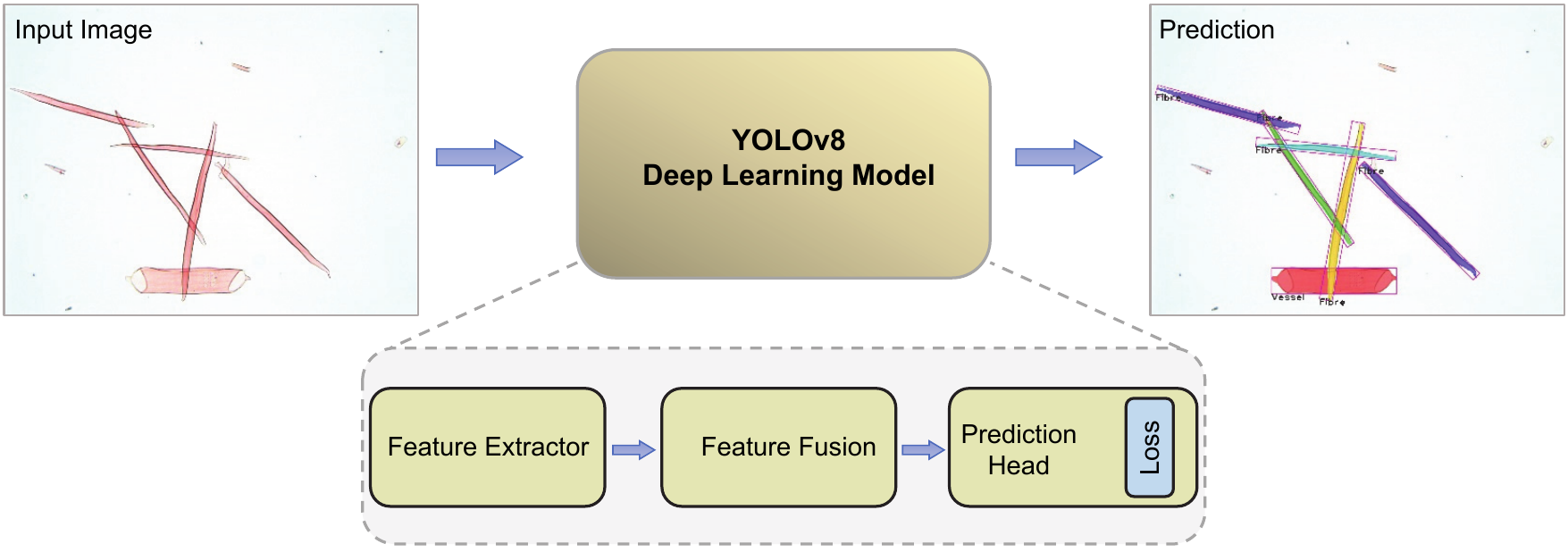}}
\caption{Schematic of the YOLOv8 architecture for fiber and vessel segmentation. The model contains a Feature Extractor for feature extraction, Feature Fusion for feature aggregation, Prediction Head for predicting the objects' bounding boxes, classes, and masks. The loss component is used to optimize the model performance. An input image is passed through the network, which performs classification, detection, and segmentation jointly. This enables the delineation of individual cells even when overlapping, as shown in the prediction output.  \label{fig1}}
\end{figure*}

While most recent light microscopes are equipped with motorized stages and the capacity to automatically generate large stitched images, the remaining limiting step is the image processing for the detection (segmentation) and shape analysis of the fibers. Several studies have developed image processing methods to analyze fibers from wood section images \cite{kennel, pan, brunel, boztoprak}. However, most of these methods are only adapted for images from cross-sections of wood samples. These samples are typically prepared with cross-section in a transverse orientation. Although this method provides crucial information about certain aspects of cell arrangement and wood density, it does not directly yield information about the length of fiber cells. Automatic segmentation of such images is also less challenging for classical image segmentation algorithms, as individual cells in the images do not overlap. As such, the image can be divided into regions (e.g., individual cells and background) where each pixel is only assigned to one cell or region. This is typically readily achievable with the classical watershed segmentation algorithm \cite{kornilov2018overview} and more recently with deep learning based segmentation tools such as Cellpose \cite{stringer2021cellpose} for more challenging samples. However, 2D images obtained from wood macerates contain many fibers that frequently overlap but which are still fully visible thanks to the translucency of the fibers imaged with the light microscope (see Fig. \ref{fig1}). Thus, many pixels in the image can belong to more than one cell of interest. This situation is challenging for most existing segmentation algorithms, including for deep learning techniques. Those are generally able to segment objects that are overlapping in nature, i.e., a human in front of a car, but can only output segmented masks that delineate the visible contours of each object and, thus, in this case, a full mask of the human in front and, an incomplete mask of the partially hidden car. With translucent objects, as is the case for light microscopy images of fibers, it would be, in principle, possible to obtain full masks for each of the overlapping objects since each object is fully visible in the image despite the overlaps. However, such an approach remains very little developed and applied \cite{bohm2018isoo, chen2022instance}.

Deep learning, a specialized subset of machine learning, has achieved significant success in fields such as computer vision and image processing, particularly in tasks like image segmentation \cite{razzak2018deep,voulodimos}. For image segmentation, convolutional neural networks (CNNs), a type of deep neural network, are well-suited. CNNs process input images through multiple filters, thereby learning autonomously features from the images without the need for humans to design feature extractors manually. CNNs are extensively used for image segmentation applications \cite{7298965, badri}, including for segmenting cells in microscope images \cite{ronneberger, fu, minaee2021image}. Given that cells in wood macerate images tend to overlap, it is essential that a neural network architecture supports the detection of individual objects, rather than merely dividing the image into regions. Instance segmentation, which identifies each cell individually and assigns pixels to the correct cell instance, is the most suitable approach in this regard. Popular existing instance segmentation methods such as Faster R-CNN \cite{ren}, Mask R-CNN \cite{he}, and RetinaNet \cite{lin} have been successfully applied to cellular image segmentation \cite{johnson, tsai, hollandi}. These methods use deep convolutional neural networks (CNNs) like VGG \cite{simonyan} and ResNet \cite{he1} to extract features from the input images. However, these two-stage methods are slow for inference. One-stage methods such as Single Shot MultiBox Detector (SSD) \cite{liu20} and You Only Look Once (YOLO) series streamline object detection by simultaneously predicting object locations and class probabilities in a single pass. One-pass approaches significantly improve inference speed over multi-stage methods. Among one-stage models, YOLO offers a good balance of speed and accuracy, achieving speeds suitable for real-time applications and that, for some cases, is more accurate than two-stage methods \cite{diwan2023object}. Most importantly, YOLOv8 can also be retrained with a setting that allows it to deal with overlapping translucent objects and thus generate a full mask of such overlapping objects. Considering our need for a versatile model that is accessible to a broad audience and capable of analyzing high-resolution stitched images of macerated fibers containing many overlapping cells, the YOLO algorithm stands out as an ideal choice for addressing this challenge.

In this paper, we develop a model based on YOLOv8 \cite{yolov8} for segmenting and classifying fibers and vessels in 2D microscopy images, see Fig. 1. We compiled a dataset of 3850 wood macerate images using 1300 microscopy images. We annotated 9 617 fibers and 519 vessels, which, after augmentation resulted in 28 358 fibers and 1502 vessels used to train the model. We demonstrate that the enhanced model achieves fast inference speeds and high accuracy in detecting and classifying individual cells when processing large images (32,640 x 25,920 pixels). We also developed a browser interface for easy model access and image analysis. This interface, accessible after local installation, enables users to upload images via drag and drop. The system then provides measurements of the fibers and vessels, including their length, width, and area, as well as the full set of segmented masks corresponding to each segmented fiber and vessel, which can be used for further detailed morphometric analysis of fiber and vessel shapes. Additionally, users can access the training and prediction code run on Google Colab \cite{colab} in the GitHub repository \cite{SaqibGithub}. Lastly, to validate the effectiveness of our new methods, we applied them to a well-known poplar transgenic line, which is recognized for having distinctly longer wood fibers than its wild-type relative. This approach aimed to corroborate previous manual measurements of fiber length differences in these lines.

\section{Image capturing method}
\subsection{Sample preparation}
To create training and test datasets, we collected stem samples from several 3-month-old hybrid aspens grown in a greenhouse, specifically from \textit{Populus tremula} L. × \textit{tremuloides} Michx.

To obtain data for comparative analysis of fiber length, 3 transgenic trees overexpressing the \textit{GA20ox1} gene Arabidopsis gibberellin 20-oxidase (Ara GA20ox1 Line 1A) \cite{eriksson2000increased} and 3 trees of the T89 clone (control) were sampled.

To perform the maceration, stem segments between internodes above 10 centimeters from the soil were used. The bark was removed from the stem and 2-3 mm of the exposed vascular cambium region was trimmed, exposing the wood. From this surface, longitudinal match-like segments of 15 mm length and approximately 1.5 x 1.5 mm cross section were prepared from the newly exposed wood surface. Maceration was performed on these match-like samples by immersing them in maceration solution (30\% Hydrogen peroxide: glacial acetic acid, 2:1 v/v) and heating at 90°C with periodic shaking for 5 hours, as previously described in \cite{siedlecka2008pectin}. The macerated solution was then sedimented by low-speed centrifugation (1000 rpm) and washed a few times with water.

To obtain training and test datasets, stem samples from several 3-month-old greenhouse-grown hybrid aspen ({\textit{Populus tremula} L. × \textit{tremuloides}} Michx.) clone T89 were collected. 
To obtain data for comparative analysis of fiber length, 3 transgenic trees over expressing the \textit{GA20ox1} gene Arabidopsis gibberellin 20-oxidase (Ara GA20ox1 Line 1A) \cite{eriksson2000increased} and 3 trees of the T89 clone (control) were sampled.

To preform the maceration, stem segments between internodes above 10 centimeters from the soil were used. The bark was removed from the stem and 2-3 mm of the exposed vascular cambium region was trimmed exposing the wood. From this surface, longitudinal match-like segments of 15 mm length approximately 1.5 x 1.5 mm cross section were prepared from the newly exposed wood surface. Maceration was performed on those match-like samples by immersing them in maceration solution (30\% Hydrogen peroxide: glacial acetic acid, 2:1 v/v) and heating at 90°C with periodic shaking for 5 hours, as previously described in \cite{siedlecka2008pectin}. The macerated solution was then sedimented by low-speed centrifugation (1000 rpm) and washed a few times with water.

For safranin staining, a few drops from the obtained macerated solution were stained with the Safranin solution (1\%) that stains lignified tissues in xylem cells. Similarly, for toluidine blue staining, a few drops from the macerated solution were stained with the (0.5\% W/V) Toluidine blue stain.

\subsection{Imaging}
The samples were mounted between a slide and coverslip and imaged using a Leica DMi8 inverted microscope in brightfield mode with transmitted white light (Leica Microsystems, Germany). The microscope is equipped with a 10X objective lens and DFC7000T color camera mounted on a 0.70X C-mount adapter. Single images were acquired in RGB color mode with a resolution of 1920 x 1440 pixels and a pixel size of 0.65 x 0.65 µm. Tile images made of 19 x 19 (361) individual images were acquired with the Navigator function of the microscope using a 10 \% overlap. Tile images were merged with the LAS X software (Leica Microsystems, Germany).

For training of the model, single tile images (1920 x 1440 pixel) of macerated fibers and vessels stained with safranin were obtained from the trimmed stem samples of wild-type (T89) trees. For quantification of the wildtype and the over-expression line, four 361-tile stitched images (containing fibers and vessels stained with safranin) were obtained from the trimmed stem samples of each of these three biological replicates for both the control and over-expression lines.

To test the robustness of the model to different staining or imaging modes we also acquired several images of fibers and vessels either stained with Toluidine blue, non stained and non stained samples acquired in grayscale mode. We provide all the raw image data at Zenodo \cite{zenodoSaqib2024}.

\subsection{Statistics}
We used a t-test statistical method to validate whether the trained model outputs consistent results across different image groups. Specifically, we tested for scale invariance by running the model on different-sized crops from the same large images. We also compared model predictions on an overexpression line GA20ox 1A and wildtype T89 as control. The t-statistic is calculated as $ t = \frac{(\bar{X}_1 - \bar{X}_2)}{\text{SED}} $,
where $\bar{X}_1$ and $\bar{X}_2$ are the sample means and $\text{SED}$ is the standard error of the difference between the means. We employed Scipy's \texttt{ttest\_ind()} to automatically compare two independent data samples, assessing significant mean differences based on t-test assumptions.

\section{Computational Method}
\subsection{Images annotation and dataset preparation}
To identify fibers and vessels in microscope images, we created a dataset of over 1300 individual images showing these structures in different shapes and sizes. The images are 1920 x 1440 pixels in size. We carefully outlined the fibers and vessels in each image using the VGG Image Annotator \cite{annot}. This created a ground truth or guide to the actual fibers and vessels in the images. The outlines were saved as JSON files, which store the coordinates of the polygons drawn around each object, which can be seen in Fig. S1. We resized the large images into smaller 1024 x 1024 pixel images to avoid running out of memory on our graphics card. We also used data augmentation techniques to increase the number of training images. For example, by applying transformations like rotating, scaling, and flipping the original images, we created more variety in the dataset. This also helps the machine learning model learn robust features that apply to new images. From 1300 original images, we generated 3850 augmented images with around 29 861 annotated objects.

To train the YOLO model, we randomly split the dataset into training (85\%) and validation (15\%) sets. The training data is used to update the model's parameters. The validation data is used to evaluate the model during training but not to update parameters. This split helps prevent overfitting and ensures the model generalizes well.

\subsection{Deep learning approach}
We used the YOLOv8-seg deep learning method, which is a variant of YOLOv8 architecture designed explicitly for instance segmentation tasks \cite{yolov8}. During the model training process, we utilized the YOLOv8 model pre-trained on the COCO val2017 dataset and set 
$overlap\_mask=False$ in training parameters to deal with overlapping masks as a starting point. The architecture of the YOLOv8 algorithm consists of four main components: feature extractor, feature fusion, prediction head, and loss function. The components are shown in Fig. \ref{fig1} and with more details in Fig. S2. Here, we look into the design concepts of each architecture module.

The feature extractor is the first part of the model and is responsible for extracting features at different stages from the input image. The output features from the different stages have different spatial resolutions. The earlier stages of the feature extractor network extract low-level features such as edges and corners. The later stages extract high-level features such as object shapes and parts. The feature extractor down-samples the input image because it extracts features at later stages.

The feature fusion module combines the output features from different stages of the feature extractor network to form a unified representation of the image. Deep neural networks capture increasingly detailed features as the network becomes deeper, which improves object prediction. However, as the network depth increases, the object localization accuracy for detecting small objects decreases owing to excessive convolution operations, resulting in the loss of important information. To address this tradeoff, the feature fusion module incorporates a multi-scale fusion of features using architectures such as a Feature Pyramid Network (FPN) and Path Aggregation Network (PAN). The feature fusion module performs different operations to extract higher-level features from the input features and consolidate the outputs from various stages of the feature extractor into a single representation. This unified representation enhances the object detection and segmentation performance.

The prediction-head module transforms the encoded image features into usable predictions for object detection and segmentation. This makes the final predictions based on the consolidated representation from the feature-fusion module. The head module combines features from earlier modules and leverages them to predict bounding boxes, classes, and masks for the objects in the image. By dividing the prediction into specialized branches, it efficiently performed classification, localization, and masking in a coordinated manner. The prediction head is the final component that outputs the actual detections and segmentations after processing using the complete YOLOv8 architecture.

The loss function in YOLOv8 measures how accurately the model detects and segments objects. It compares the model predictions with the ground-truth labels. The loss function is used to train the model to improve its performance. YOLOv8 has separate branches for classification, bounding-box regression, and masking. For classification and masking, the cross-entropy loss was used to minimize errors. Bounding box detection uses two losses: distributed focal loss (DFL) and CIoU Loss. These consider the aspect ratio between the predicted and ground-truth boxes. The overall loss is the sum of the losses from different branches. A lower loss indicates that the model is more accurate at detecting, classifying, and segmenting objects. Loss guides model training to improve these areas.

\subsection{Measuring the morphology of objects}
To identify and localize objects of interest within images, we use the YOLOv8 model to predict bounding boxes and contours around each instance. To enable more precise morphological measurement, we integrate functions from the OpenCV computer vision library to further analyze the object contours \cite{opencv}.

We use the contours to generate the mask of each predicted object by the model. We then calculate the length of the skeleton path of the binary mask. For this, we apply a thinning operation that is used to reduce the thickness of objects in an image to a single-pixel-wide skeleton. The goal of thinning is to preserve the essential structural and topological characteristics of the original objects while significantly reducing the amount of data. To apply the thinning process, we use the OpenCV library function cv2.ximgproc.thinning(). We then analyze the skeletons in the thinned image, extract the long path skeleton, and calculate its length using np.sum(mask) from the NumPy library. This centerline approach allows us to find the actual length of the object.

To calculate the width of the fiber, the code first computes the distance transform of the binary mask drawing. The distance transform assigns each pixel a value that corresponds to the distance from that pixel to the nearest background pixel. This is done using the cv2.distanceTransform() function. The maximum value in the distance transform represents the thickness of the thickest part of the fiber. This value is obtained using the np.max() function from the NumPy library. Since the distance transform gives the distance from the center of the fiber to the edge, the actual thickness of the fiber is twice this value.

\begin{figure}[htb]
\centering{\includegraphics[clip, trim=0cm 0cm 0cm 0cm, width=0.9\linewidth]{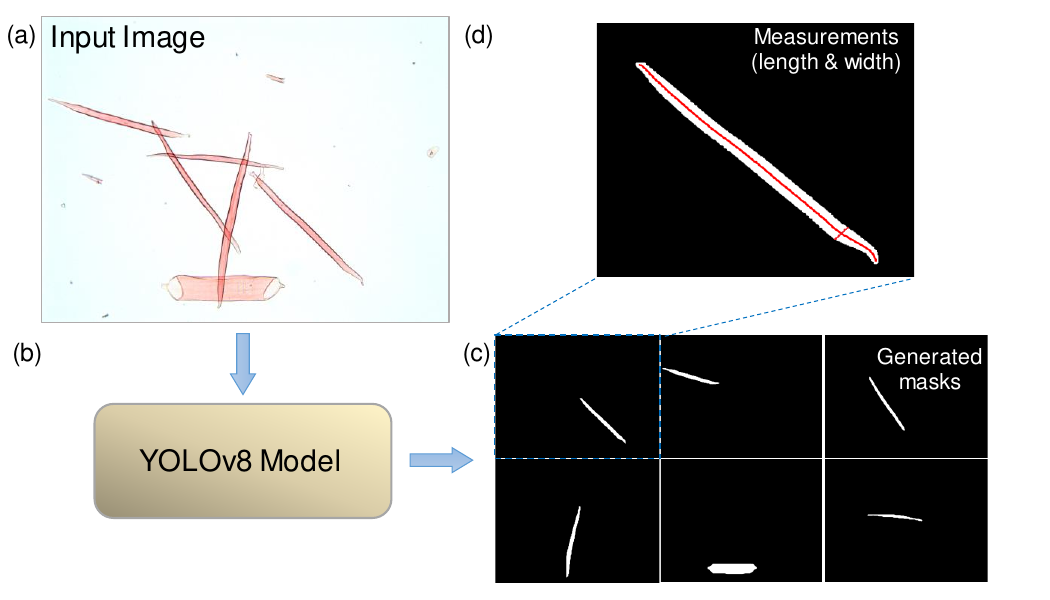}}
\caption{Example of a segmented object and the corresponding image analysis that automatically measures the object's morphological traits using our YOLOv8 model. Panel (a) displays the original microscopy input image used for object detection, which is sent to the YOLOv8 model (b) that outputs detected objects. Panel (c) shows the individual mask generated for each individual cell in the image and also demonstrates that full masks are obtained from translucent overlapping objects. Panel (d) illustrates the measurements of length and width for each detected object, providing quantitative data extracted from the image. \label{fig2}}
\end{figure}

In addition to length and width, we used OpenCV's cv2.contourArea() method to accurately measure the area enclosed within the detected contours. A schematic diagram showing the concept is shown in Fig. \ref{fig2}, where the input image is passed to the model and the model gives the output. From the output, we generate masks for each instance and then use OpenCV functions to measure the length and width of each instance. The output from the model provides pixel values from the analysis, which need to be converted into micrometers.

By utilizing the polygon points in the contours, we perform post-processing to exclude segmented objects that touch the boundary edge of the image. This post-processing approach ensures that only fully segmented objects within the image are included.

\section{Model retraining and parameter adjustment}
\subsection{YOLOv8 model retraining and hyperparameter tuning with wood macerate dataset}
During the training process of all models, we utilized the pre-trained YOLOv8 models on the COCO val2017 dataset as a starting point. To train the YOLO model, we use 3350 images for training and 500 images for validation. The training data is used to update the model's parameters whereas the validation data is used to evaluate the model during training but not to update parameters. We used YOLOv8-m, l, and x in our experiment. YOLOv8-m uses a medium-sized feature extractor and more feature fusion levels, YOLOv8-l utilizes a larger feature extractor and more feature fusion levels compared to YOLOv8-m, and YOLOv8-x uses an even larger feature extractor and more feature fusion compared to YOLOv8-l. We find the convergence level and best optimizer for these models during training. Based on experimental data from Ultralytics, we observed that YOLOv5 training required 300 epochs, while YOLOv8 training increased the number of epochs to 600. Initially, we set the number of epochs to 600 and incorporated a patience value of 50. This means that if no noticeable improvement occurred after waiting for 50 epochs, the training would terminate early. However, during the training of YOLOv8m, we found that the model reached its best performance at epoch 355 and training stopped early at epoch 405. 

We chose hyperparameters for model training as suggested in reference \cite{yolov5}. The selection of an appropriate optimization algorithm is crucial when training YOLOv8-seg. The optimizer determines how the model parameters are updated during training to minimize the loss function. For small custom datasets, the Adam (Adaptive Moment Estimation) optimizer is recommended, while the SGD (Stochastic Gradient Descent) optimizer tends to perform better on larger datasets. Consequently, we trained YOLOv8-seg models separately using the Adam and SGD optimizers. The results of comparing the effects of these two optimizers on model training are presented in Table \ref{tab:1}. 

\begin{table*}[h]
\centering
\caption{The table compares YOLOv8m-seg, YOLOv8l-seg, and YOLOv8x-seg models trained on our generated dataset using both SGD and Adam optimizers.}
\label{tab:1}
\vspace{5pt}
\resizebox{0.85\textwidth}{!}{%
\begin{tabular}{|l|l|l|l|l|l|}
\hline
\textbf{Model} & \textbf{Optimizer} & \textbf{Best Epoch} & \textbf{mAP 50} & \textbf{\begin{tabular}[c]{@{}l@{}}mAP 50-95\end{tabular}} &
  \textbf{\begin{tabular}[c]{@{}l@{}}Speed\\ GPU V100      (ms)\end{tabular}} \\
\hline
YOLOv8m-seg    & SGD   & 405       & 0.93         & 0.73   & 8.3     \\
               & Adam    & 413     & 0.94         & 0.75    & 9.0        \\
\hline
YOLOv8l-seg    & SGD    & 421     & 0.93         & 0.74  & 17.4       \\
               & Adam   & 487    & 0.94         & 0.76  & 17.2      \\
\hline
YOLOv8x-seg    & SGD   & 553      & 0.93         & 0.76  & 24.5     \\
               & Adam  & 543     & 0.94         & 0.78   & 22.2       \\
\hline
\end{tabular}%
}
\end{table*}

We opted for the Adam optimizer with a weight decay of 5x$10^{-4}$ and an initial learning rate of 1x$10^{-3}$. Furthermore, we set the input image size to 1024 and trained the different models using a TITAN V100 16GB with a batch size of 8. The models were trained on Python 3.8 and PyTorch 1.10.0.

\subsection{Evaluation metrics for the models}
To evaluate the performance of the models, we used four metrics: precision, recall, F1-score, and mean average precision (mAP). These metrics are commonly used for object detection and segmentation tasks, and they are calculated based on the number of true positive (TP), true negative (TN), false positive (FP), and false negative (FN) predictions made by the model \cite{sokolova2009systematic}.

Precision quantifies how many of the predicted positive instances are actually correct. Recall quantifies how many of the actual positive instances are correctly identified by the model. F1-score combines both precision and recall giving a single value that represents the algorithm's overall accuracy. A higher F1-score indicates a better algorithm performance in achieving both high precision and high recall simultaneously.

Mean Average Precision (mAP) is a widely adopted evaluation metric for assessing the performance of object detection algorithms across multiple classes. In this paper, we considered mAP 50 and mAP 50-95, in which, mAP 50 calculates the average precision for all classes at an IoU threshold of 0.5 while mAP 50-95 computes the average precision for all classes over a range of IoU thresholds from 0.5 to 0.95, with a step size of 0.05. This variation of mAP offers a more comprehensive evaluation by considering a wider range of IoU thresholds.

\section{Result and Discussion}
\subsection{Model Selection}
In this work, we used the pre-trained YOLOv8 for detecting and segmenting fibers and vessels. We trained our custom dataset using three YOLO variants: m, l, and x. The quantitative results for precision, recall, F1 score, mAP@0.5, and mAP@0.5-0.95 values of the three YOLOv8 models in fiber and vessel segmentation are presented in Table \ref{tab:2}. Among these models, YOLOv8m-seg demonstrated the highest precision (0.97), recall (0.91), and F1 score (0.94), while performing relatively lower in mAP: 0.5-0.95 (0.75). YOLOv8l-seg exhibited slightly lower precision (0.96) and recall (0.91) compared to YOLOv8m-seg, but had a comparable F1 score (0.93) and a marginally better mAP: 0.5-0.95 (0.76). YOLOv8x-seg matched YOLOv8m-seg in precision (0.97), had a slightly lower recall (0.90), and an F1 score (0.93) equal to YOLOv8l-seg, but it outperformed both in mAP: 0.5-0.95 with the highest value of 0.78. Furthermore, the YOLOv8x-seg model had the highest weight (70.1), compared to YOLOv8m-seg (27.2) and YOLOv8l-seg (45.9).

Based on these findings, the YOLOv8x-seg model was identified as the optimal choice for fiber and vessel detection and segmentation, exhibiting superior performance across the evaluation metric of mAP@0.5-0.95. Consequently, the YOLOv8x-seg model was selected for further analysis, specifically in estimating fiber and vessel length, width, and area. This selection ensures a consistent and focused evaluation of the model's practical application, aligning with the study's objectives.

\begin{table*}[htb]
\centering
\caption{The table shows the evaluation metrics for different YOLOv8 models when detecting fibers and vessels.}
\label{tab:2}
\vspace{5pt}
\resizebox{0.9\textwidth}{!}{%
\begin{tabular}{|l|l|l|l|l|l|l|}
\hline
\textbf{Model} & \textbf{Precision} & \textbf{Recall} & \textbf{F1-Score} & \textbf{mAP 50} & \textbf{mAP 50-95} & \textbf{\begin{tabular}[c]{@{}l@{}} Weight \end{tabular}} \\
\hline
YOLOv8m-seg    & \textbf{0.97}  & \textbf{0.91} & \textbf{0.94} & 0.95         & 0.75           & 27.2              \\
\hline
YOLOv8l-seg    & 0.96           & 0.91          & 0.93           & 0.94          & 0.76            & 45.9              \\
\hline
YOLOv8x-seg    & 0.97           & 0.90          & 0.93           & \textbf{0.94} & \textbf{0.78}   & 70.1              \\
\hline
\end{tabular}%
}
\end{table*}

\subsection{Model Performance}
We thoroughly evaluated model performance for the tasks of detecting and segmenting fiber and vessel objects in images. We examined precision-recall curves to understand the trade-off between precision and recall for detection. For segmentation, we focused on how precisely the model delineates the boundaries and segments of the detected objects. Additionally, we analyzed F1-confidence curves to understand the relationship between F1 scores and model confidence levels for detection and segmentation. Examining the F1-confidence curves provided insights into how precision and recall were balanced across varying confidence thresholds. Evaluating the model's performance step-by-step is important for assessing its effectiveness.

\begin{figure*}[htb]
\centering{\includegraphics[clip, trim=0cm 0cm 0cm 0cm, width=1.0\linewidth]{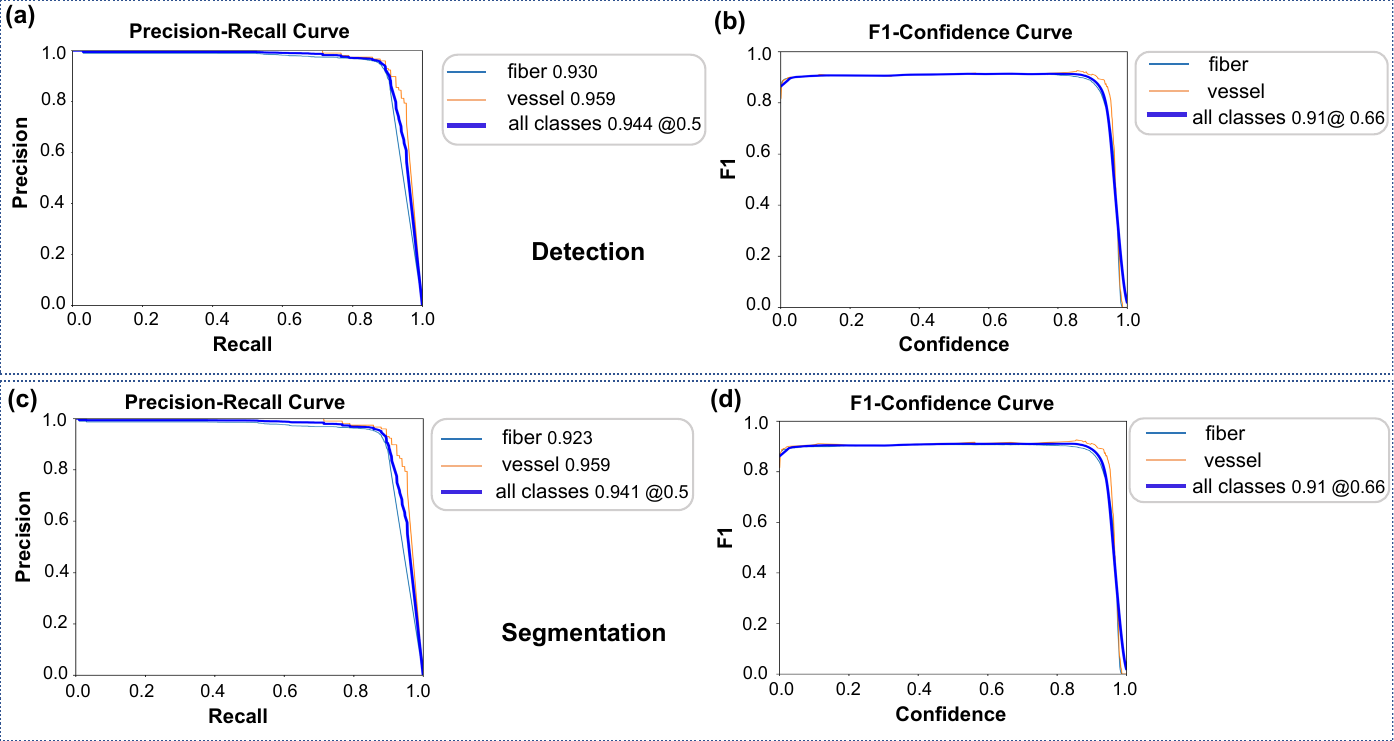}}
\caption{These plots showcase the precision-recall (a, c) and F1-confidence (b, d) curves used to evaluate the performance of YOLOv8x in detecting (a,b) and segmenting (c,d) fibers and vessels. The model demonstrates strong mAP and F1-score across thresholds, confirming its effectiveness in object detection and segmentation tasks. \label{fig3}}
\end{figure*}

Figure \ref{fig3} depicts the behavior of the selected YOLOv8x-seg model used for object detection and segmentation of fiber and vessel objects. The precision-recall curve in Fig. \ref{fig3}(a) represents the trade-off between precision and recall for the detection task. At a threshold of 0.5, the mean average precision (mAP) values were 0.930 for fiber and 0.959 for vessel detection. The overall mAP of 0.944 for all classes combined indicated the model's overall performance in object detection.

Figure \ref{fig3}(b) shows the F1-confidence curve, which illustrates the relationship between the F1 score and the model's confidence. At a confidence threshold of 0.66, the F1-score was 0.91 for both fiber and vessel classes. This score represents the balance between precision and recall. A higher F1-score indicates better model performance in terms of both precision and recall. These findings provide valuable insights into the model's performance and help assess its suitability for detecting fiber and vessel objects in images.

Moving on to the segmentation task, the model achieved values of 0.923 and 0.959 at a threshold of 0.5 for fiber and vessel segmentation, respectively. The overall mAP of 0.94 for all combined classes at the same threshold is shown in Fig. \ref{fig3}(c), highlighting the model's effectiveness in segmenting objects. Figure \ref{fig3}(d) displays the F1-confidence curve for the segmentation task, where the model attained an F1-score of 0.91 at a confidence threshold of 0.66. This score represents the trade-off between precision and recall for both fiber and vessel classes.

We conclude that for fiber and vessel detection and segmentation, the YOLOv8x-seg model performed best compared to other models in terms of mAP. The YOLOv8m-seg model had the lowest mAP value specifically for fiber detection and segmentation. 

\subsection{Qualitative Results}
\begin{figure*}[htb]
\centering{\includegraphics[clip, trim=0cm 0cm 0cm 0cm, width=1.0\linewidth]{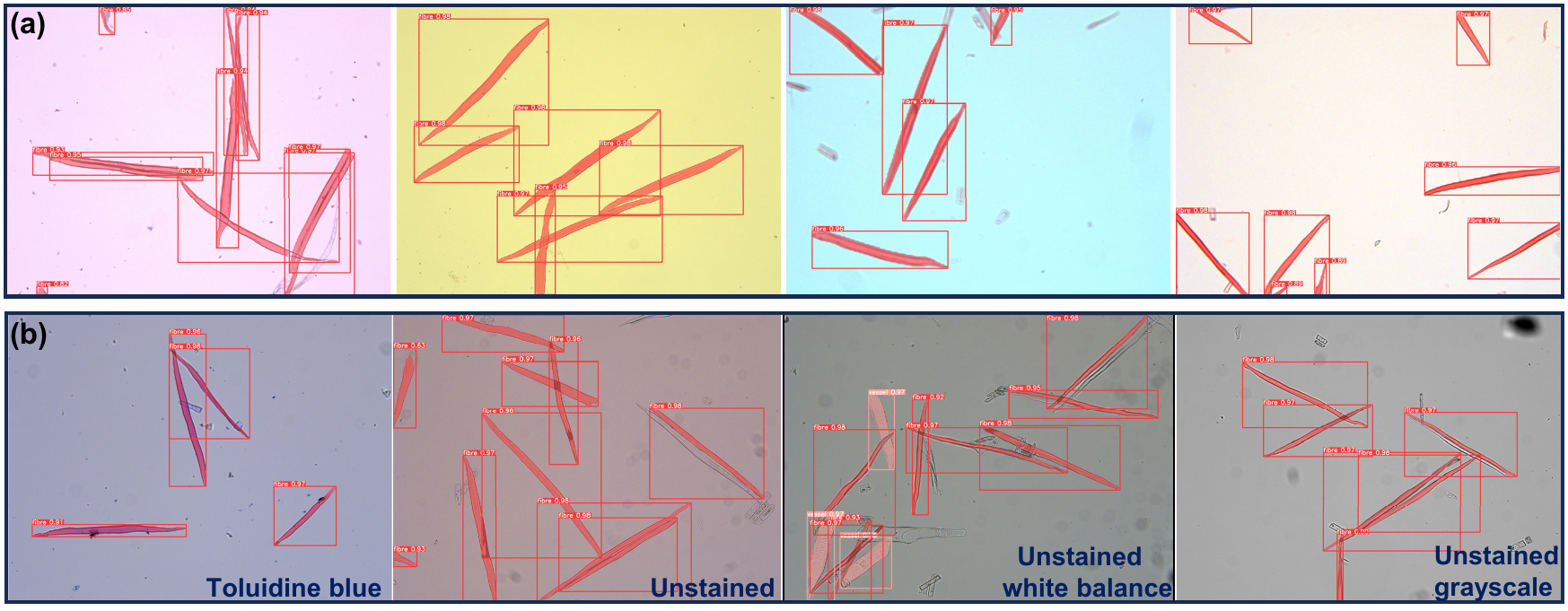}}
\caption{The images depict YOLOv8 effectively detecting and segmenting images obtained with various staining protocols and color acquisition parameters. Panels in (a) depict only Safranin stained fibers with diverse backgrounds typically associated with imperfect white balance and (b) shows from left to right, Toluidine blue stained samples, unstained, unstained with properly adjusted white balance, and unstained acquired in grayscale mode. Note that in these representations, the overlayed mask of fibers and vessels is displayed as red and should not be confused with red staining. \label{fig4}}
\end{figure*}

Figure \ref{fig4} shows fiber and vessel detection and segmentation examples achieved using the YOLOv8x-seg model. These examples highlight the model's ability to accurately identify and outline fiber and vessel structures in the images. The visual results obtained from the model not only demonstrate its potential within the scope of our study but also provide valuable insight into its practical application for fiber and vessel analysis.

Because the model was primarily trained on RGB color images of safranin stained samples, we next aimed to test whether it was also able to detect and segment fibers in images obtained with different staining protocols and color acquisition parameters. We acquired several images with imperfect white balance figure \ref{fig4}(a), Toluidine blue stained samples, unstained samples, and unstained samples acquired in grayscale mode (all other images were acquired in RGB color mode; figure \ref{fig4}(b). Despite these variations, our model consistently succeeded in detecting and segmenting fiber and vessel structures. This highlights the effectiveness of the model, as it still can perform well. Such robustness is particularly valuable in fiber and vessel segmentation, where images with diverse backgrounds are encountered.

\begin{figure}[htb]
\centering{\includegraphics[clip, trim=0cm 0cm 0cm 0cm, width=1.0\linewidth]{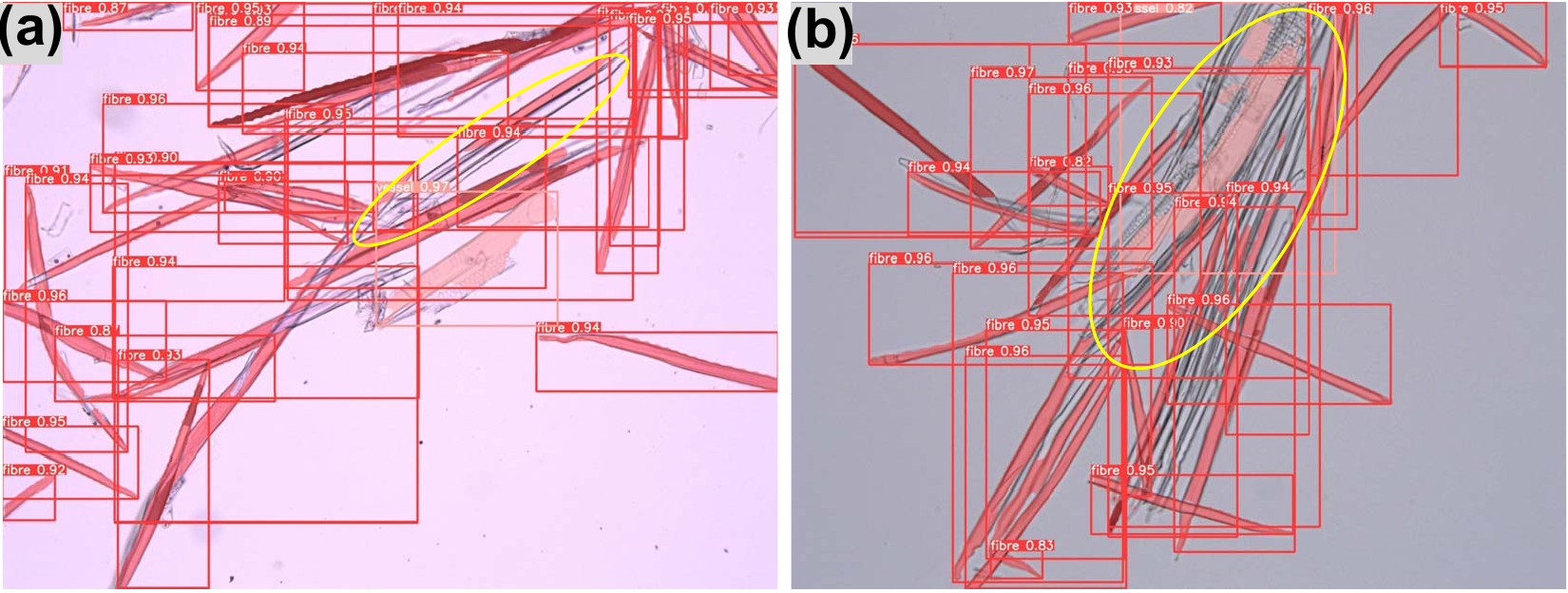}}
\caption{The detection of fibers and vessels encountered some challenges in densely packed regions, as shown in these examples where several fibers and vessels were not properly detected or segmented. \label{fig5}}
\end{figure}

Based on the findings of this study, we conclude that our retrained version of the YOLOv8x-seg model effectively detects and segments fibers and vessels in microscopy images. Despite achieving high accuracy in fiber detection and segmentation, our YOLOv8 model still made some mistakes by generating false positives and false negatives in certain cases. We highlight examples of this in Fig. \ref{fig5} (a), where the model failed to identify several fibers and fiber segments, for instance, highlighted by the yellow ellipse. Similarly, in Fig. \ref{fig5}(b), the ellipse highlights one of the regions where YOLOv8 failed to identify the fiber and vessel. This failure to detect the fiber could be attributed to limitations in the training of the model but is also likely a consequence of the very high density and crowding of the fibers in this example image. This is an issue that can also be easily alleviated during sample preparation by increasing the dilution of the sample before mounting it between the slide and coverslip and imaging. Nevertheless, to tackle this issue from the computational side, Adar et al. \cite{adar} suggested that training the model with a larger dataset containing more input features can greatly enhance its ability to generalize and perform well on new and unseen data. A larger dataset would enable the model to capture the subtle differences in fiber and vessel structures in various background images. Additionally, a larger dataset can help to mitigate the risk of overfitting, where the model becomes too specialized to the training dataset and performs poorly on new samples.

\subsection{Quantifying fibers and vessel in microscopy images}
In order to complete the task of fiber and vessel detection and segmentation, we implemented standard measurements to describe the shape of the objects based on the model output. Specifically, we evaluated how well our model performed with images of different sizes. We separated the images into two categories based on size. The first type measured 1920 x 1440 pixels, similar to those used to train the model. The second type was large tile images of 33,384 x 25,112 pixels stitched from 361 small 1920 x 1440 images. The model first identified the locations of fibers and vessels in the image. Then, it assesses the morphological traits (e.g., length, width and area) of the detected objects using the method described in subsection 3.3. The total detected items in two large images are 1818 fibers and 62 vessels. Note that we implemented a post-processing step that removes fibers that are touching the borders in the image to avoid a bias of shorter fibers due to partially segmented objects.

\begin{figure*}[htb]
\centering{\includegraphics[clip, trim=0cm 0cm 0cm 0cm, width=1.0\linewidth]{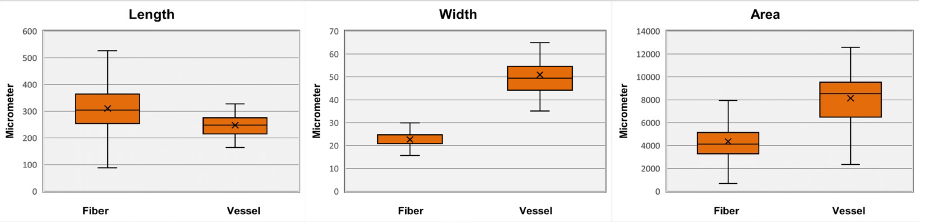}}
\caption{Boxplots depict fibers' and vessels' length, width, and area measurements from two large images of 33,384 x 25,112 pixels. The box plots show the 1st and 3rd quartiles (box limits), mean values (cross sign), and median values (solid line).  \label{fig6}}
\end{figure*}

Figure \ref{fig6} shows box plots summarizing the measurements of the detected fibers and vessels. The fibers' length ranged from 254-364 µm (average 310 µm), the width ranged from 21-25 µm (average 23 µm), and the area ranged from 3 283-5 145 µ$m^2$ (average 4 347 µ$m^2$). Whereas for most vessels, the length ranged from 218-275 µm (average 247 µm), the width ranged from 44-55 µm (average 51 µm), and the area ranged from 6 536-9 494 µ$m^2$ (average 8 142 µ$m^2$). In the supplementary materials, scattered and histogram plots offer a detailed visualization of the distribution and relationship between the length, width, and area as depicted in Fig. S3.

Further, testing on different image sizes allowed us to check if the model performance stayed consistent regardless of size changes. The results show that the model's effectiveness is independent of image dimensions. This scale-invariance makes the model more robust and can analyze images of varying sizes which are encountered in real applications. This property is crucial for real-world problems, as different image dimensions will inevitably be encountered in practice. To evaluate scale invariance and object detection and segmentation abilities, including overlapped objects, 20 small images of size 1920 x 1440 pixels were randomly selected, containing 115 fibers and 17 vessels. Measurements were compared to those from the large 33,384 x 25,112 pixel image.

We found the model extracted consistent average dimensions for fibers across these two image sizes. This includes an average length of 376 µm, a width of 23 µm, and an area of 10 689 µ$m^2$ for the fibers. For fiber length, width, and area, the p-values were 0.954, 0.963, and 0.945, respectively, indicating no statistically significant difference between small and large image measurements. However, we noted a variability when comparing the vessel measurements with a lower average length of 269 microns and an average width of 57 microns. The vessel area averages were closer between image sizes at 17 755 µ$m^2$ on small images and 13 533 µ$m^2$ on large images. The p-values for vessel length, width, and area were 0.275, 0.142, and 0.454, respectively. These results illustrate the scale invariance performance of the model across different image dimensions. In the supplementary material, we also provide the analysis of 20 small-sized images (1920 x 1440), see Fig. S4 and two mid-sized images consisting of 99 tiles (8275 x 7250), see Fig. S5, using box plots, scatter plots, and histogram plots to assess the robustness of the model.

In conclusion, the model's ability to accurately extract metrics for fibers and vessels, including those that overlap, at various scales, underscores its proficiency in managing overlapping objects and scale invariance. Such capabilities are crucial for the robust detection and segmentation of high-resolution images, ensuring consistent performance regardless of image size. This model, therefore, becomes a valuable tool for research groups aiming to quantify metrics from extensive datasets.

\subsection{Comparison of GA20ox 1A line and wildtype T89}
To further evaluate our approach for research applications, we tested it on a new dataset consisting of 12 large images of samples taken from T89 wildtype trees and 12 large images from the transgenic line overexpressing the \textit{GA20ox1} gene. This transgenic line was previously reported to show approximately 10 \% increase in fiber length compared to the wildtype T89 \cite{eriksson2000increased}. In total, the model identified 5717 fibers in the 12 T89 images with an average length of 422.5 µm and for the GA20 line it identified 6303 fibers with an average of 471.25 µm, see Fig. 7. A student t-test indicated that the samples were highly significantly different, as evidenced by a p-value less than 0.0001. These results show an approximately 12 \% length increase for the GA20ox line compared to the wildtype, in line with previously reported results \cite{eriksson2000increased}. Overall, these results suggest that our new high-throughput microscopy-based AI-assisted fiber characterization method works accurately to quantify differences between wild-type and mutant lines. As such it will be particularly useful in the future to study much larger mutant and natural variation tree collections.

\begin{figure*}[htb]
\centering{\includegraphics[clip, trim=0cm 0cm 0cm 0cm, width=1.0\linewidth]{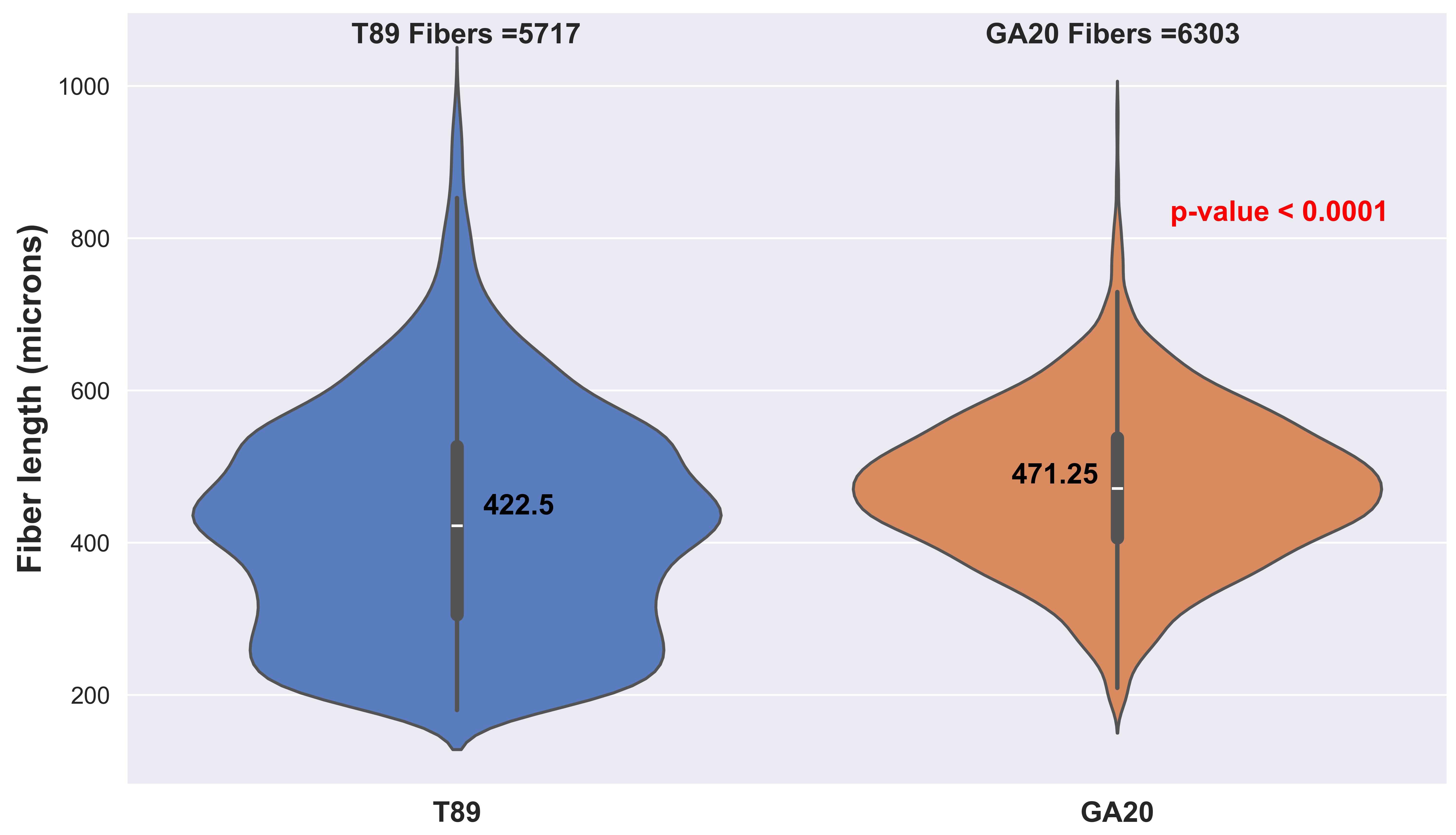}}
\caption{Violin plots that depict fibers' length measurements from large images of 33,384 x 25,112 pixels from T89 and GA20ox line. The envelope shows the distribution, the thick center lines represent the 1st and 3rd quartiles (black line limits), and median values (center yellow dots). \label{fig7}}
\end{figure*}

\subsection{GUI application and GitHub resources}
We have developed a desktop application using Python’s Flask framework, designed to simplify the process of uploading and analyzing images with our automated algorithm. The application's workflow is shown in Fig. S6. Users can select an image from a gallery or upload one via drag-and-drop. Note, that the image should be in a RGB format. Once an image is uploaded, the application predicts the presence of fibers and vessels. It also quantifies each detected object, with results available in a downloadable data file. Note that segmented objects that touch the edge of the image and may thus not be fully segmentable, are automatically excluded from the analysis at this postprocessing step. This file includes details like object type, length, width, and area. It also generates a folder containing the full set of segmented masks corresponding to each segmented fiber and vessel. Such binary masks can then be further analysed to quantify other shape descriptors of interest using, for example, OpenCV \cite{opencv}, \cite{van2014scikit}, and ImageJ/Fiji \cite{abramoff2004image,schindelin2012fiji,legland2016morpholibj}. When installed on a local computer with an i9 10th Gen processor and 32GB of RAM, the application processes a 1024 x 1024 image in an average of 140 milliseconds.

Detailed instructions for installation of the developed code and instructions for implementation are found in the supporting information. Additionally, users can find instructions on how to retrain and test the model using Google colab in the GitHub resources \cite{SaqibGithub}. The 'Train$\_$custom$\_$data.ipynb' file can be used to retrain the model; simply follow the instructions provided in the file. The 'prediction$\_$file.ipynb' file is used to test the model on your custom dataset.

\section{Conclusion}
This paper introduces a deep learning solution using YOLOv8 to automatically analyze and quantify wood fibers and vessels in challenging microscope images, offering high-throughput capabilities. To achieve this, we trained multiple YOLOv8 models on diverse wood image datasets and evaluated their performance in detecting and segmenting fibers and vessels. The most robust model was chosen. The model can consistently and reliably extract essential cell metrics across different image scales, such as length, width, and area. The model's consistent metric extraction underscores its strong practical applicability. We also created a web application pipeline that is useful in practical situations. Users can then upload images for automatic cell counting and shape quantification. Thus, we conclude that this study introduces an innovative high-throughput method for analyzing wood cells in densely populated 2D microscopy images, even when cells are partially obscured.
\backmatter

\bmhead{Supplementary information}
Additional information and user guides are included in the supporting information file. 

\section*{Declarations}
\subsection*{Funding}
The project was funded by Kempestiftelserna (JCK-2129.3) and the National Academic Infrastructure for Supercomputing in Sweden (NAISS) at Umeå, partially funded by Vetenskasrådet (2022-06725). The authors acknowledge the facilities and technical assistance of the Umeå Plant Science Centre Microscopy facility and the plant growth facility. This work was also supported by grants from the  Wallenberg Foundation (KAW 2016.0341 and KAW 2016.0352), VINNOVA (2016–00504), and the Novo Nordisk Foundation (NNF21OC0067282) to S.V. We also thank Bio4Energy for supporting this work.

\subsection*{Declaration of competing interest}
The authors declare that they have no known competing financial interests or personal relationships that could have appeared to influence the work reported in this paper.

\subsection*{Consent for publication}
The authors declare that they have no competing interests.

\subsection*{Availability of data and code}
All the code in this project was developed using Python and different public libraries as defined in the supporting information. The Google Colab project is found here \cite{colab} and the code can be downloaded at a public GitHub repository here \cite{SaqibGithub}. All raw image data are available at Zenodo\cite{zenodoSaqib2024}.

\subsection*{Authors' contributions}
Saqib Qamar: Data curation, Software, Methodology, Validation, Formal analysis. Abu Imran Baba: Investigation, Validation. Stéphane Verger: Conceptualization, Resources, Supervision, Funding acquisition. Magnus Andersson: Conceptualization, Resources, Supervision, Funding acquisition, Project administration. All authors contributed to Writing, review and editing.

\subsection*{Ethics approval and consent to participate}
Not applicable.





\bibliography{references}

\end{document}